\documentclass[]{gmaart2019}

\usepackage[utf8]{inputenc} 
\usepackage[T1]{fontenc} 
 
\usepackage{amsmath,amssymb,amsfonts,mathtools,amsthm} 

\usepackage[babel,autostyle]{csquotes}

\usepackage{tabularx}
\newcolumntype{L}[1]{>{\raggedright\arraybackslash}p{#1}} 
\newcolumntype{C}[1]{>{\centering\arraybackslash}p{#1}} 
\newcolumntype{R}[1]{>{\raggedleft\arraybackslash}p{#1}} 
\usepackage{booktabs}

\usepackage{paralist}   

\usepackage[noend]{algpseudocode}

\algrenewcommand\algorithmicrequire{\textbf{Voraussetzung:}}
\algrenewcommand\algorithmicensure{\textbf{Abschlussbedingung:}}

\usepackage{listings} 
\usepackage{subcaption}  

\usepackage{epstopdf}
\usepackage{xcolor}
\selectcolormodel{gray}  

\usepackage{scrhack}
\usepackage{varwidth}
\usepackage{rotating} 
\usepackage[defaultlines=2,all]{nowidow}  
\usepackage{import}
\usepackage{placeins}


\usepackage{glossaries}
\usepackage{xspace}
\usepackage{colortbl}
\usepackage{moresize}
\usepackage{siunitx}
\usepackage{adjustbox}
\usepackage{pgfplots}
\usepackage[super]{nth}

\begin{document}


\hyphenpenalty=2000

\pagenumbering{roman}
\setcounter{page}{1}
\pagestyle{scrheadings}
\pagenumbering{arabic}

\setnowidow[2]
\setnoclub[2]

\newcommand\stress[1]{{\color{red}#1}}
\newcommand\ie{i.\,e.\xspace}
\newcommand\eg{e.\,g.\xspace}
\newcommand\etal{et\,al.\xspace}

\selectcolormodel{natural}

\addto\extrasenglish{
    \renewcommand{\chapterautorefname}{Chapter}
    \renewcommand{\sectionautorefname}{Section}
    \renewcommand{\subsectionautorefname}{Subsection}
    \renewcommand{\subsubsectionautorefname}{Subsection}
    \renewcommand{\paragraphautorefname}{Paragraph}
}

\makeatletter
\newcommand\footnoteref[1]{\protected@xdef\@thefnmark{\ref{#1}}\@footnotemark}
\makeatother


\renewcommand{\Title}{Concepts for Automated Machine Learning\\in Smart Grid Applications}


\renewcommand{\Authors}{Stefan Meisenbacher, Janik Pinter, Tim Martin,\\Veit Hagenmeyer, Ralf Mikut}
\renewcommand{\Affiliations}{\footnotesize{
        Institute for Automation and Applied Informatics,
        Karlsruhe Institute of Technology\\
		Hermann-von-Helmholtz-Platz 1,
		76344 Eggenstein-Leopoldshafen\\
		E-Mail: stefan.meisenbacher@kit.edu}}
							 
\renewcommand{\AuthorsTOC}{S.~Meisenbacher, J.~Pinter, T.~Martin, V.~Hagenmeyer, R.~Mikut} 
\renewcommand{\AffiliationsTOC}{Karlsruhe Institute of Technology} 

\setLanguageEnglish
							 
\setupPaper 



\section{Introduction}
\label{sec:Introduction}

Undoubtedly, the increase of available data and competitive machine learning algorithms has boosted the popularity of data-driven modeling in energy systems.
Applications are forecasts for renewable energy generation \cite{Ahmed2020, Qian2019} and energy consumption \cite{Nti2020}.
Forecasts for load and generation, \eg, power, gas, and heat, on different temporal and spatial aggregation levels are elementary for sector coupling, where energy-consuming sectors are interconnected with the power-generating sector to address electricity storage challenges by adding flexibility to the power system \cite{Fridgen2020}.
However, the large-scale application of machine learning algorithms in energy systems is impaired by the need for expert knowledge, which covers machine learning expertise and a profound understanding of the application's process.
The process knowledge is required for the problem formalization, as well as the model validation and application.
The machine learning skills include the processing steps of i) data pre-processing, ii) feature engineering, iii) algorithm selection, iv) HyperParameter Optimization (HPO), and possibly v) post-processing of the model's output.
\smallbreak
\begin{figure}
\centering
    \includegraphics[width=1\linewidth]{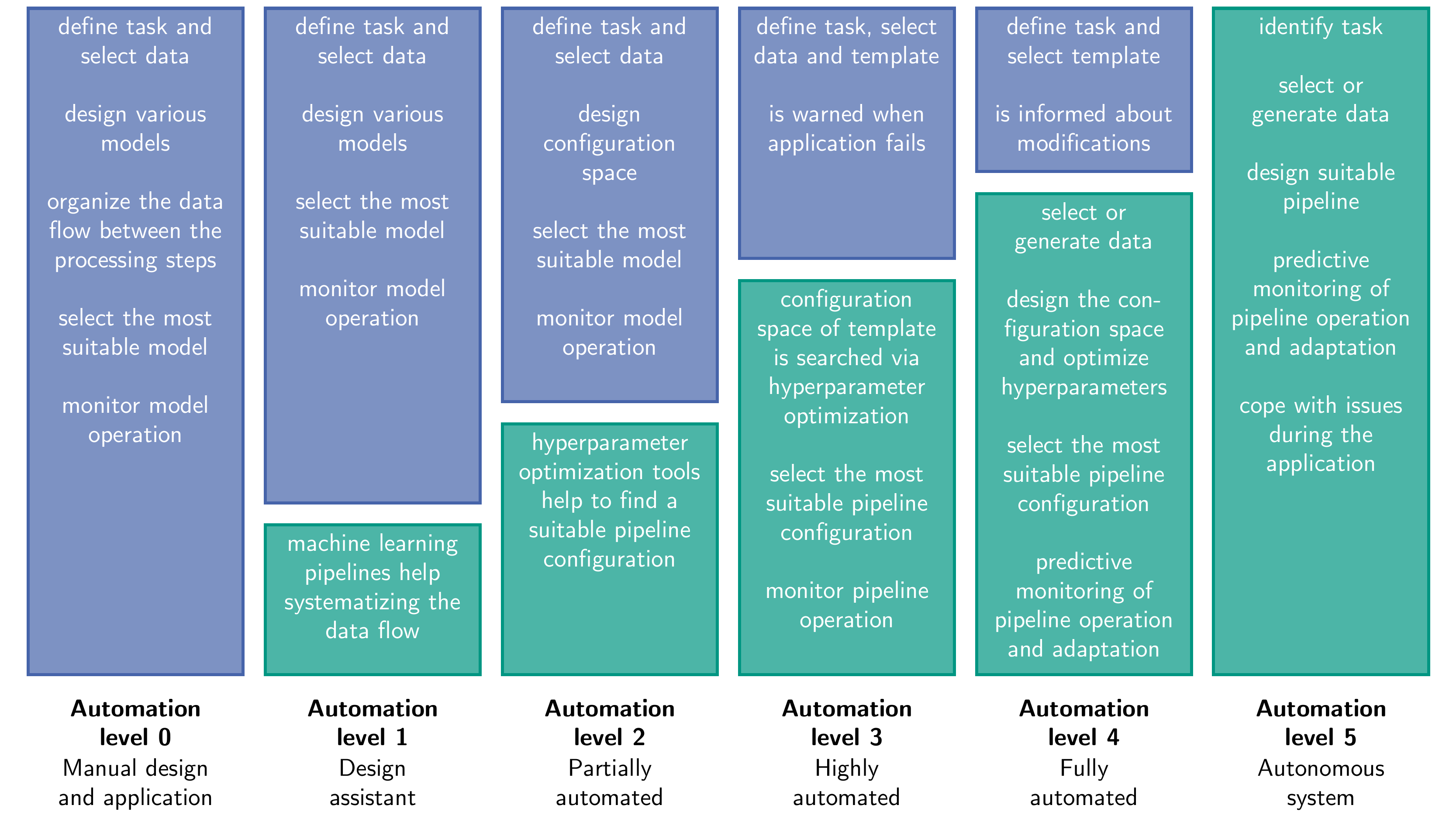}
    \caption{The five levels of automated forecasting, inspired by the SAE standard for autonomous driving of vehicles \cite{SAE-J3016}.}
    \label{fig:automation_levels}
\end{figure}
Tailoring a model for a particular application requires selecting the data,
    designing various candidate models and organizing the data flow between the processing steps,
    selecting the most suitable model,
    and monitoring the model during operation -- an iterative and time-consuming procedure.
Automated design and operation of machine learning aim to reduce the human effort to address the increasing demand for data-driven models.
We define five levels of automation for forecasting where manual design and application reflect \textbf{Automation level 0}, see \autoref{fig:automation_levels}.
\smallbreak
In \textbf{Automation level 1}, machine learning pipelines \cite{Heidrich2021} assist the design process, systematizing the workflow by serially organizing the processing steps and managing the data flow through the steps’ methods.
Still, the pipeline requires manual tailoring by the data scientist to meet the specific requirements.
Most published literature on energy forecasts range between Automation level 0 and 1, see reference \cite{GonzalezOrdiano2018}.
Across the literature, standard procedures have emerged that can be used to create automated pipeline templates for specific tasks.

Partially automated forecasting is enabled in \textbf{Automation level 2}, where HPO tools\footnote{
    \eg, Hyperopt \cite{Hyperopt}, SMAC \cite{SMAC3}, or NNI \cite{NNI}
} support the data scientist, automatically evaluating candidate models of a configuration space $\boldsymbol{\Lambda}$ defined by the data scientist.
Still, the data scientist needs to analyze the optimization results, select the most suitable model, and monitor the model during operation.
In the literature, few approaches exist for energy systems that we can classify as Automation level 2.
A framework for automated HPO and forecasting algorithm selection is proposed by Rätz \etal \cite{Ratz2019}.
Cui \etal \cite{Cui2016}, and Shahoud \etal \cite{Shahoud2020a} propose frameworks for the automated forecasting algorithm selection using meta information such as statistical properties of the time series and characteristics of the system.
An approach for combining HPO and ensembling of forecasting algorithms is proposed by Wu \etal \cite{Wu2020}.
Maldonado \etal \cite{Maldonado2019} and Valente and Maldonado \cite{Valente2020} introduce embedded feature selection approaches for the Support Vector Regression (SVR), integrating exogenous weather information into electrical load forecasting.

\textbf{Automation level 3} reaches highly automated forecasting by providing pipeline templates for specific tasks that include an associated configuration space $\boldsymbol{\Lambda}$ or a robust default configuration $\boldsymbol{\lambda}$.
The data scientist needs to provide the data and select the template.
Anomaly detection monitors operation and alerts the data scientist when suspicious model inputs or outputs are detected.\footnote{
    Current open-source Automated Machine Learning (AutoML) tools, \eg, AutoSklearn \cite{AutoSklearn}, or TPOT \cite{TPOT}, support automated design of regression and classification models. Monitoring of the model operation is not provided.
}
A highly automated framework for building energy management is proposed by Schachinger \etal \cite{Schachinger2018}, including a heuristic for the automated design of Artificial Neural Networks (ANNs), online assessment, and scheduled re-training.

In \textbf{Automation level 4}, the fully automated forecasting takes over the data selection.
The data is either taken from a data storage assigned to the selected template or generated synthetically according to the template-specific task.
During operation, the model predicts its performance and warns the data scientist before system borders are reached.

Finally, \textbf{Automation level 5} achieves a fully autonomous system that independently identifies the task, creates the model, and detects and resolves issues during operation.

The introduced automation levels are not rigid -- interim levels are possible.
To the best of our knowledge, there are yet no applications for smart grids in Automation levels 4 and 5.
\smallbreak
The remainder of this paper is organized as follows.
First, we present a general approach to automate the design and operation of forecasting models in energy systems in \autoref{sec:approach}.
Then, we describe and evaluate automated design algorithms for a hybrid model (autonomous level 2.5) in \autoref{sec:HybridModeling}.
Finally, \autoref{sec:conclusion_outlook} concludes and provides an outlook on future research.

\section{Approach}
\label{sec:approach}

\begin{figure}
\centering
    \includegraphics[width=1\linewidth]{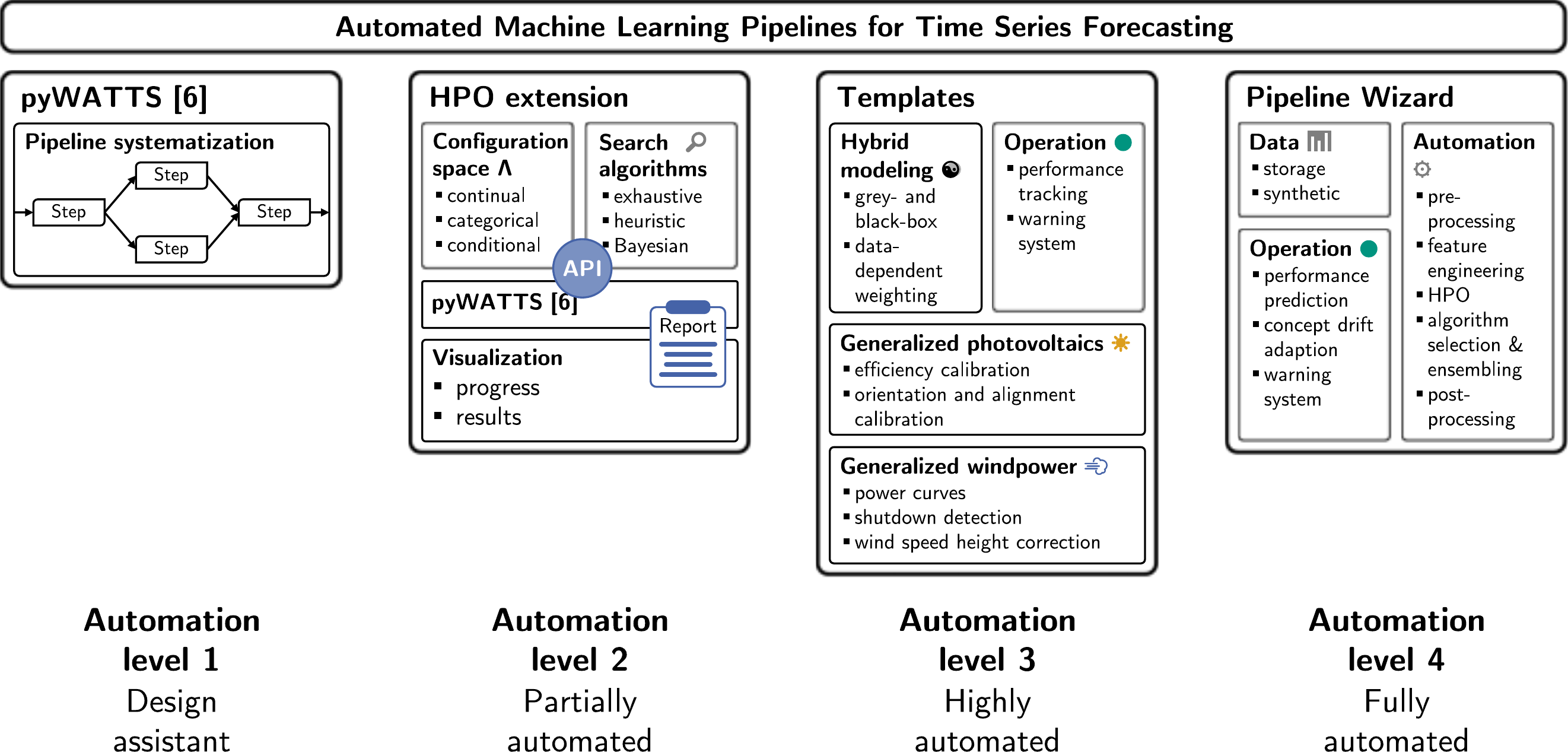}
    \caption{Identified Automated Machine Learning (AutoML) approaches for smart grid application. The HyperParameter Optimization (HPO) extension communicates via the Application Programming Interface (API) with pyWATTS \cite{Heidrich2021}.}
    \label{fig:automation_approach}
\end{figure}

Although numerous methods for AutoML have been proposed in the literature, a toolkit tailored for forecasting in energy systems is lacking.
\autoref{fig:automation_approach} shows a schematic overview of unexplored automation approaches and our long-term concept based on a taxonomy discussed in \autoref{ssec:LiteratureReview}.  
\smallbreak
\begin{sloppypar}
The open-source Python Workflow Automation Tool for Time Series (pyWATTS)\footnote{
    \url{https://github.com/KIT-IAI/pyWATTS}
} \cite{Heidrich2021} assists researchers in the design process, systematizing the workflow through a pipeline with a uniform interface for various methods applied to the steps of the pipeline (\textbf{Automation level 1}).
\end{sloppypar}

For specialized tasks, the expertise of a data scientist and a process expert is still required, and we want to keep the human in the loop.
To reduce the effort of tailored pipeline design, an HPO extension for pyWATTS \cite{Heidrich2021} is required (\textbf{Automation level 2}).
The extension enables defining a configuration space $\boldsymbol{\Lambda}$ and selecting a search algorithm, and wraps around pyWATTS.
Communication is established by an Application Communication Interface (API) of pyWATTS, allowing the optimization algorithm to configure pipeline parameters.
The report interface of pyWATTS provides data for visualization of the optimization progress and results.
The schematic process of HPO with pyWATTS is outlined in \autoref{ssec:HPOExtension}.

Recurring tasks with good generalizability, such as forecasting of PhotoVoltaic (PV) and Wind Power (WP) generation, can be handled with default templates for large-scale deployment (\textbf{Automation level 3}).
A default template contains a forecasting pipeline with normalized output that needs little effort to calibrate for new operational environments.
We introduce a template for PV forecasting in \autoref{ssec:PhotovoltaicsTemplate}, and a template for WP forecasting in \autoref{ssec:WindPowerTemplate}.
\\
The hybrid modeling template couples two grey- or black-box models by a data-dependent weighting of the model outputs.
In regions where the model input is well represented in the training data set, a sophisticated model is overweighted,
    whereas, in less representative regions, a robust model gains weight, as it is expected to have better extrapolation characteristics.
We evaluate exhaustive and Bayesian HPO for the automated design of a black-box hybrid model without operation monitoring (\textbf{Automation level 2.5}) on ten benchmark data sets in \autoref{sec:HybridModeling}.
\\
The operation of the templates is supported by performance monitoring and a warning system, alerting the data scientist if any issue is detected during operation, such as unusually high forecasting errors.
\smallbreak
The vision for fully automated pipeline design for energy systems requires a tool specific to energy systems to integrate domain knowledge -- the \textit{Pipeline Wizard} (\textbf{Automation level 4}).
For tasks where comprehensive training data is missing, the data manager automatically selects appropriate training data from a related data storage or synthetically generates training data.
The \textit{Pipeline Wizard} automates the design of the forecasting pipeline and enables integrating specific methods for each pipeline section.
The operation of the \textit{Pipeline Wizard} is guided by predictive performance estimation to detect drifts in the pipeline error at an early stage.
This is required to trigger automated model adaption to cope with concept drifts and informs the data scientist about changes made.
Detailed information on the planned realization of the \textit{Pipeline Wizard} can be found in \autoref{ssec:PipelineWizard}.

\subsection{Literature Review}
\label{ssec:LiteratureReview}
Review papers are fundamental for the evaluation of the state of science and the identification of research gaps.
In the research area of AutoML, several literature review papers exist, \eg, \cite{Zoeller2021, Santu2021}.
However, they are limited to regression and classification tasks.
Further, AutoML methods focus on the problem of Combined Algorithm Selection and Hyperparameter optimization (CASH) \cite{Zoeller2021}.
For time series forecasting, pre-processing and feature engineering are vital sections of the machine learning pipeline and require specialized methods, considering the temporal sequence of data points.
Consequently, a review on AutoML for time series forecasting must consider time series-specific methods and the complete pipeline -- an unaddressed issue in the present review studies.

\subsection{Hyperparameter Optimization Extension for pyWATTS}
\label{ssec:HPOExtension}

\begin{figure}
\centering
    \includegraphics[width=1\linewidth]{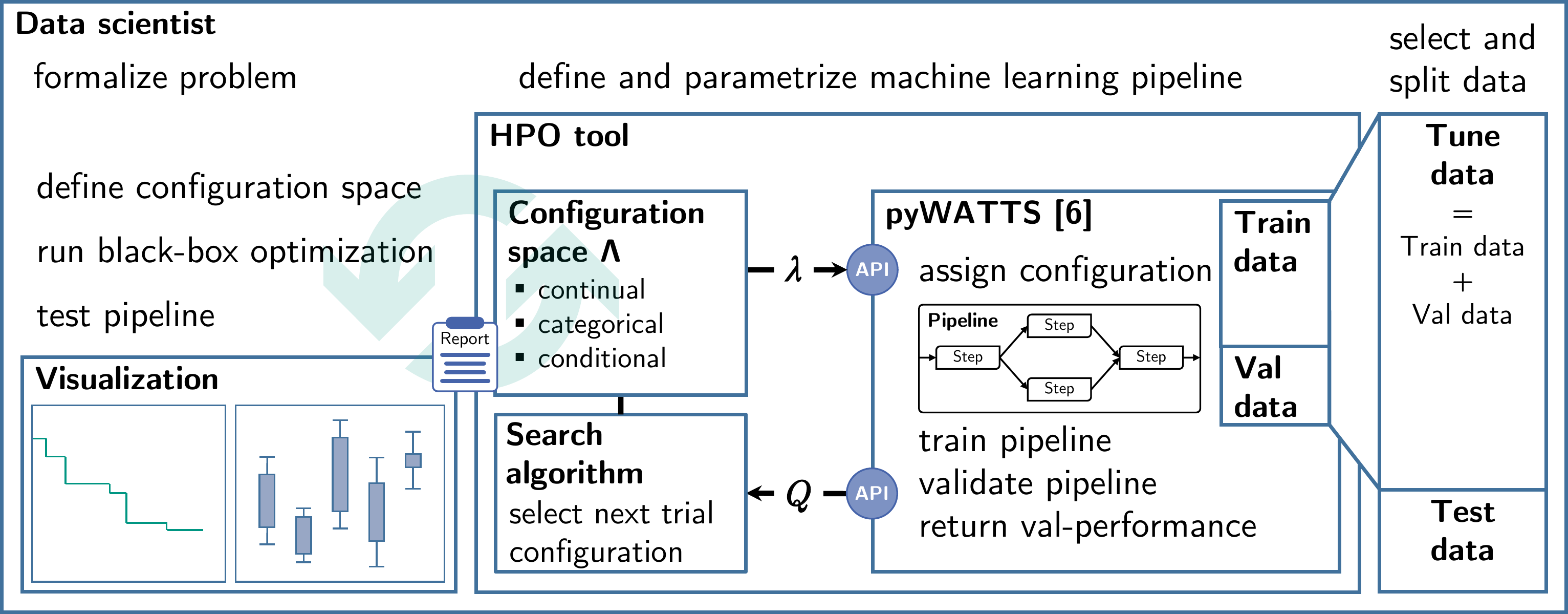}
    \caption{The schematic process of HyperParameter Optimization (HPO) with the pyWATTS extension: The data scientist is supported by the HPO tool in tailoring the pipeline to a specific problem. The HPO tool passes a hyperparameter configuration $\boldsymbol{\lambda} \in \boldsymbol{\Lambda}$ via the Application Programming Interface (API) to pyWATTS \cite{Heidrich2021} and receives the pipeline performance $Q$ on the validation (val) data.}
    \label{fig:HPO_extension}
\end{figure}

Systematizing the workflow with machine learning pipelines can be achieved with pyWATTS  \cite{Heidrich2021}.
For enabling external HPO tools to access the pipeline configuration $\boldsymbol{\lambda}$ of pyWATTS, we target to define an API.
\smallbreak
The schematic process for HPO is shown in \autoref{fig:HPO_extension}.
The data scientist formalizes the problem, defines the structure of the machine learning pipeline, and selects the data.
The data needs to be split into a set for tuning and a test set.
The tuning set is used to find a suitable pipeline configuration $\boldsymbol{\lambda}$.
We further split the tuning data set and use a portion to train the pipeline and evaluate the performance of $\boldsymbol{\lambda}$ on the validation set.\footnote{
    To increase the robustness, Cross-Validation (CV) can be applied.
}
The test set is hold-out to evaluate the tuned pipeline afterward.
\\
The data scientist parametrizes the pipeline sections to be optimized and defines the configuration space $\boldsymbol{\Lambda}$ accordingly.
For the definition of $\boldsymbol{\Lambda}$, continuous, categorical, and conditional terms are available.
While continuous terms are used to define the configuration space of hyperparameters, categorical terms are used for making decisions, such as choosing a polynomial or Radial Basis Function (RBF) kernel of an SVR or selecting an ANN or SVR as the forecasting algorithm.
Depending on the choice, conditional terms enable the definition of corresponding sub-configuration spaces $\boldsymbol{\Lambda}_\text{cond} \subset \boldsymbol{\Lambda}$, \eg, the degree of the polynomial kernel if this kernel was selected.\footnote{
    The applied definition of a configuration space $\boldsymbol{\Lambda}$ is shown in \autoref{ssec:Evaluation}.
}

The HPO tool selects a hyperparameter configuration $\boldsymbol{\lambda} \in \boldsymbol{\Lambda}$, which is assigned to the pipeline.
pyWATTS trains and validates the pipeline and returns the performance $Q$ on the validation data split, which is usually the forecasting error.
Depending on the selected search algorithm of the HPO tool, $Q$ is used for the selection of the next $\boldsymbol{\lambda}$ to be evaluated or not.
We target to implement the open-source HPO tool Neural Network Intelligence (NNI) \cite{NNI}, which allows the selection of various search algorithms, including exhaustive, heuristic, and Bayesian algorithms, while the definition of the configuration space $\boldsymbol{\Lambda}$ is standardized.
\\
In HPO, parallel computing is crucial for feasible run times.
We may parallelize the pipeline's training process, the computation of CV folds, the computation of configuration trials, or combinations of these.
The best parallelization strategy depends on the actual problem and can be determined in a preceding experiment.
The HPO extension for pyWATTS will include the above strategies.
\\
The evolution of the pipeline performance during optimization and the optimization results need to be visualized to aid the data scientist in the design process.
The evolution plot of $Q$ may indicate the convergence of directed search algorithms and guide the data scientist in deciding on the termination.
Visualization of the best performing hyperparameter configurations helps the developer to decide whether $\boldsymbol{\Lambda}$ was well defined.

\subsection{Generalized Photovoltaics Template}
\label{ssec:PhotovoltaicsTemplate}

\begin{figure}[t!]
\centering
    \includegraphics[width=1\linewidth]{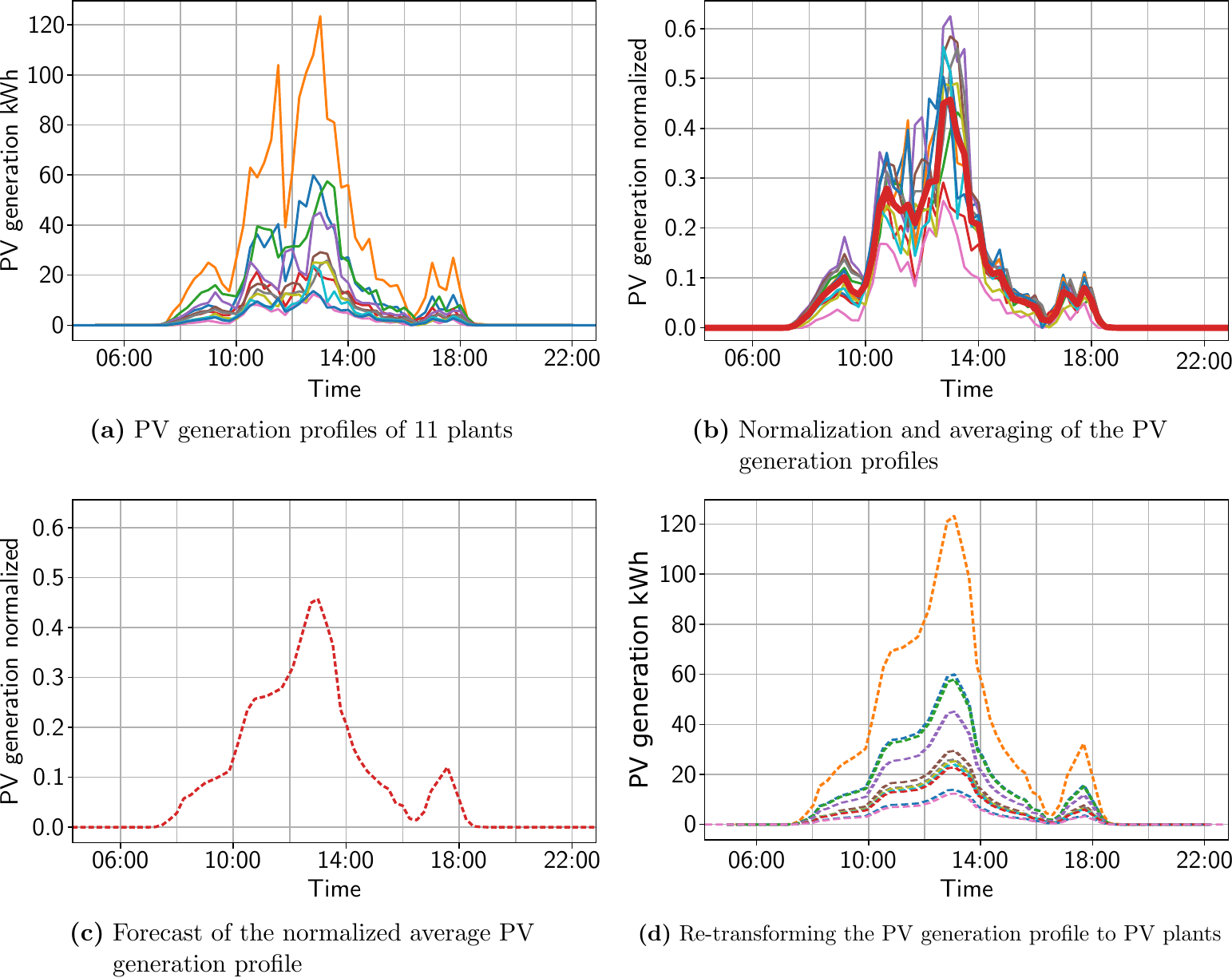}
    \caption{The process of the generalized PhotoVoltaic (PV) generation forecasting template.}
    \label{fig:PV-template}
\end{figure}

The majority of published literature on PV forecasting is limited to individual plants, \eg, \cite{Chaouachi2010, Mellit2010, Atique2019}.
They differ in terms of input features, forecasting horizon, and forecasting algorithms.
The increasing adoption of renewable energies and their integration into redispatch policies leads to a rapidly growing demand for PV forecasting models.
Therefore, we expect that designing and training an individual model for each PV plant is infeasible due to the immense design effort and the need for a sufficient amount of training data for each plant, which are not present for new plants.
Several commercial solutions exist for renewable energy forecasting in the context of redispatch actions, \eg, \cite{GridSage, FuturePowerFlow, Redispatching}.
However, the applied methods are closed-source, making an evaluation in terms of forecasting performance and design efficiency impossible.
\smallbreak
We propose a generalized PV forecasting template, which uses weather forecasts for the plant's location as input data -- more precisely, global radiation and air temperature \cite{Martin2021}.
Thus, weather forecasting is an external module for which we may use a commercial weather forecasting service or an individual weather model.
\autoref{fig:PV-template} shows the process of the generalized PV generation forecasting template.
We generalize the model using normalized training data of eleven PV plants, whose alignment and orientation are either unknown or ambiguous.
After normalizing the generation profiles according to the peak power of the PV plants, we calculate the average generation profile.
We train the generalized PV template to forecast the average normalized generation profile with the weather forecast as input data.
After forecasting the average normalized profile, we re-transform the generation profile to individual plants in the post-processing.

\begin{figure}
\centering
    \includegraphics[width=1\linewidth]{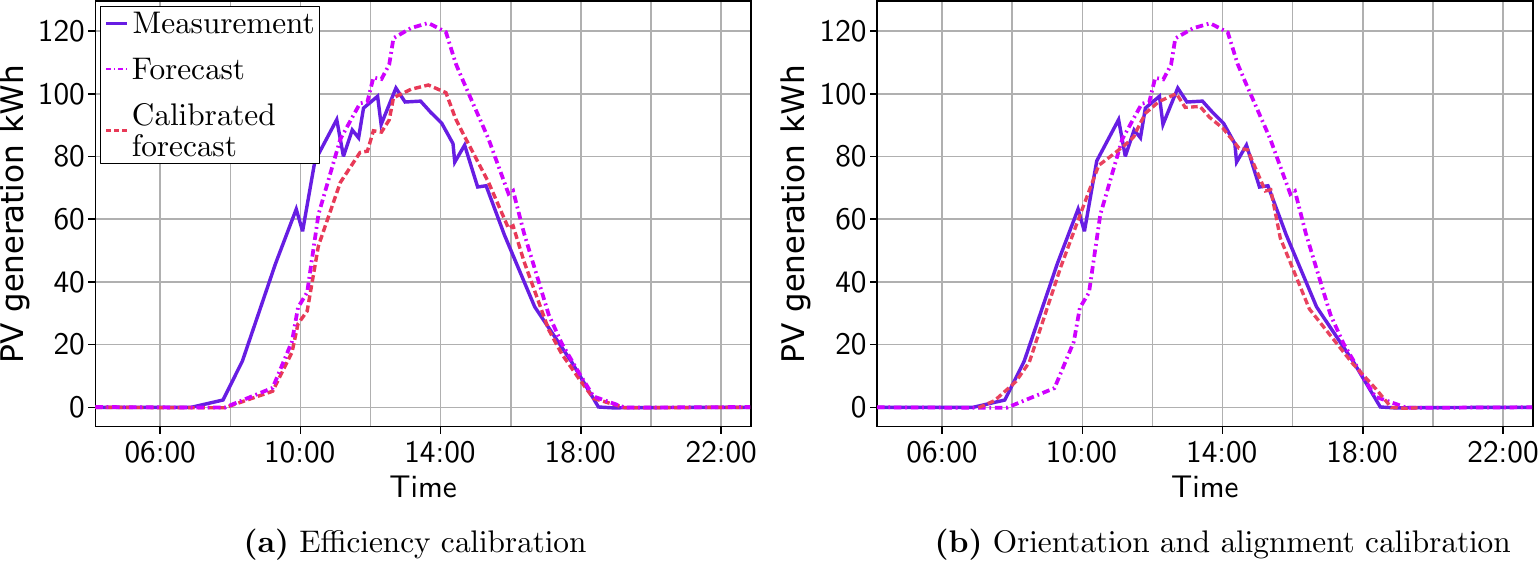}
    \caption{The calibration of the generalized PhotoVoltaic (PV) generation forecasting template.}
    \label{fig:PV-calibration}
\end{figure}

We validate this approach out-of-sample and achieve a normalized Mean Absolute Error (nMAE) of $\SI{26.3}{\percent}$.
We may reduce the nMAE to $\SI{15.9}{\percent}$ if we would use a flawless weather forecast. 
To reduce the forecasting error of the template, it seems reasonable to introduce calibration factors.
The factors allow the calibration of the template to individual plants to compensate for different efficiency levels, as well as inclinations and orientations, see \autoref{fig:PV-calibration}.
In addition, the PV template will support various complex forecasting models depending on the availability of data, \ie, depending on the availability and amount of site-specific historical data and weather forecasts.

The proposed generalized PV model was developed for the \textit{Stadtwerke Karlsruhe Netzservice GmbH}.
For the automated application, we target to implement online performance tracking and calibration.
Thereby, the model can adapt to decreasing efficiencies due to aging or changing environmental conditions, \eg, shading from new buildings in the surrounding area.
Once re-calibration is performed, the data scientist is informed about the modification.

\subsection{Generalized Wind Power Template}
\label{ssec:WindPowerTemplate}

\begin{figure}
\centering
    \includegraphics[width=1\linewidth]{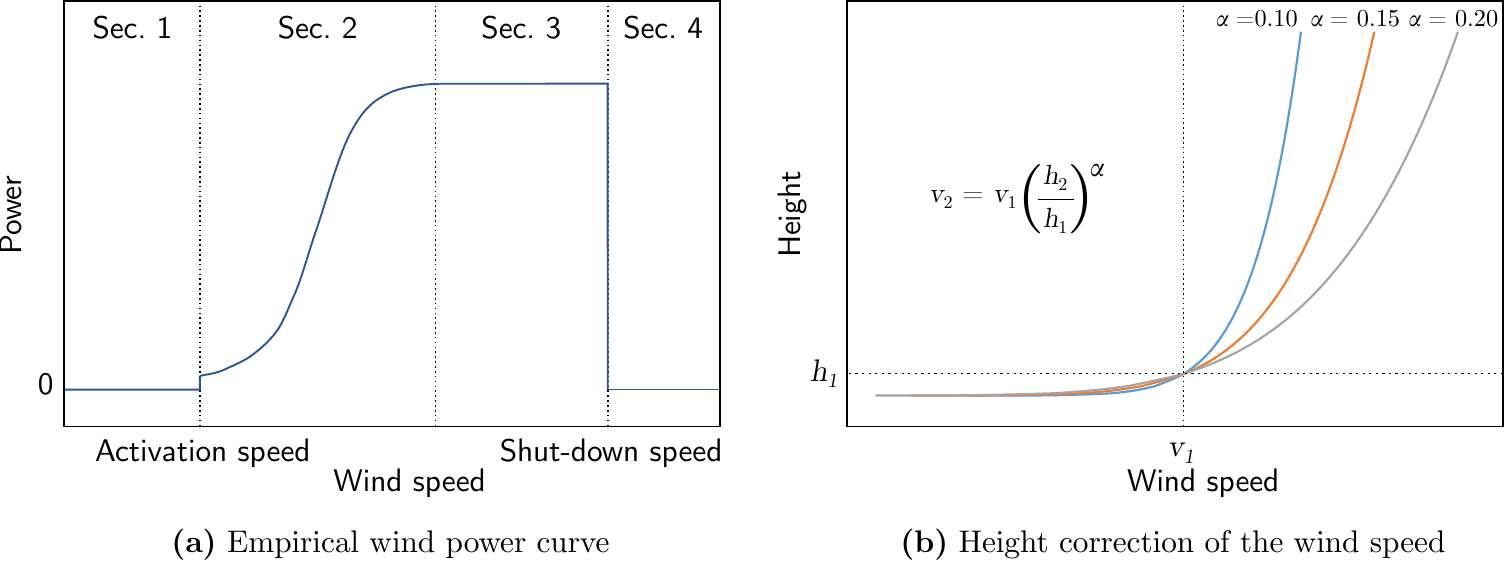}
    \caption{The wind power curve and height correction with the wind profile power law for reference height $h_1$ and the wind speed at this height $v_1$ with different exponents $\alpha$.}
    \label{fig:WP-curve}
\end{figure}

Wind turbine manufacturers provide empirical power curves, which link the power output of the wind turbine to the wind speed at hub height.
We propose to use these power curves to forecast WP generation, rather than designing and training individual data-driven models \cite{Martin2021}.
The input of a power curve is the wind speed of a weather forecast, which comes from a commercial service or an individual weather model.
As the wind speed of the weather forecast is not at hub height, a correction is necessary.
We use the wind profile power law
\begin{equation}
    \frac{v_{2}}{v_{1}}=\left(\frac{h_{2}}{h_{1}}\right)^{\alpha},
\end{equation}
where $v_1$ and $v_2$ are the wind speeds at height $h_1$ and $h_2$ above the ground, and $\alpha$ is the empirically derived friction coefficient, depending on the topology of the landscape \cite{Masters2004}.
\autoref{fig:WP-curve} shows the four sections of a power curve and the height correction with the wind profile power law.
Using the height correction, we are able to calibrate the power curve to the respective turbine.
In most cases, the heights $h_1$ and $h_2$ are known.
The velocity $v_1$ is the wind speed of the weather forecast.
In this case, we calibrate the power curve with the exponent $\alpha$.
In the literature, $\alpha$ is given for different landscape topologies, which serve as a starting value for the calibration.
The utilization of calibrated empirical power curves eliminates the need for extensive training data.
If training data is available, re-calibration of $\alpha$ is possible.
In the first application, we determined $\alpha$ using grid search and selected $\alpha$ such that the forecasting error on a validation data set becomes minimal.\footnote{
    If apart from $\alpha$, $h_1$ or $h_2$ is unknown, calibration is still possible but the effort increases.
}

We validate the calibrated power curve out-of-sample and achieve an nMAE of $\SI{62.6}{\percent}$.
The model error seems high, but it is mainly related to the forecasting accuracy of the wind speed.
A flawless forecast of the wind speed reduces the nMAE to $\SI{19.6}{\percent}$.
The wind speed forecast, in particular, has difficulty predicting single wind gusts,
    justifying the large discrepancy between the nMAE obtained with weather forecasts and flawless forecasts.
During the compilation of the validation data set, we noticed anomalies due to the manual shutdown of the wind turbines.
To clean the data set, we used a heuristic method that takes advantage of the fact that there are two turbines in the immediate neighborhood and searches for deviating outputs between the turbines.
In further work, we target to develop a universal anomaly detection method that also works for individual turbines.

As the PV template (\autoref{ssec:PhotovoltaicsTemplate}), the WP template was developed for the \textit{Stadtwerke Karlsruhe Netzservice GmbH}.
We aim to implement online performance tracking and calibration, as well as data-dependent complexity of the forecasting model.
In this way, the model may adapt to changing inflow characteristics, \eg, caused by transformations of the landscape topology.
Re-calibration triggers the information of the data scientist about the model adaptation.

\subsection{Pipeline Wizard}
\label{ssec:PipelineWizard}

The objective of the \textit{Pipeline Wizard} is the automated forecasting pipeline design and large-scale application for consumption data in energy systems, including electricity, gas, and heat.
For these systems, sufficient training data is not always available.
However, we may use data from related systems for training with similar environmental conditions, unit size, and utilization.
The related data either can be taken from a data warehouse or generated synthetically.
In order to select or generate suitable data, meta information is necessary that describe the behavior of the system.
By eliminating the need for measurement data of the system, forecasting models can be applied widely.

Recent AutoML tools focus on the CASH problem for classification and regression tasks.
Time series forecasting requires specialized methods for pre-processing and feature engineering that consider the temporal sequence of the data.
We target to provide time series-specific methods for each section of the pipeline, including pre-processing, feature engineering, HPO, algorithm selection, and ensembling.
Apart from these default methods, integrating specialized methods for particular system domains is possible, which are then taken into account in the automated design process, \eg, copy-paste imputation for energy time series to handle anomalies \cite{Weber2021} or the engineering of energy-specific meta-features \cite{Cui2016}.

Concept drifts pose a major challenge in the application of forecasting models.
A concept drift involves the change of the target variable's statistical properties \cite{Zliobaite2016}.
Reasons for concept drifts are manifold: a change of utilization, changing exogenous influences affecting the system, or structural changes such as unit size or system boundaries.
The changes may occur suddenly, incrementally, or gradually and may reoccur \cite{Gama2014}.
At the same time, the forecasting accuracy of a model decreases if the trained relationships between input and output variables no longer match the system's behavior.
In this situation, adapting the forecasting model to the changed system is necessary \cite{Zliobaite2016}.
Different adaptation strategies are possible.
The most straightforward strategy is re-training the forecasting model with the data accumulated after the concept drift.
An improvement can be achieved if not only the model parameters but also its structure and hyperparameters are tuned.
Celik and Vanschoren \cite{Celik2021} evaluated six adaptation and tuning strategies on evolving data for a classification task.
We target to tailor these strategies to time series forecasting and evaluate their effectiveness.

\section{Automated Hybrid Modeling}
\label{sec:HybridModeling}

Robust models are necessary for the representation of participants in smart grids, \eg, the thermal dynamics of buildings or the characteristics of Electric Vehicle (EV) batteries \cite{Schwenk2021}.
Data-driven models can achieve high predictive accuracy if the input variables are in familiar range, thus, similar to the training data set (\textit{interpolation)}.
The prediction accuracy declines in \textit{extrapolation} areas \cite{Boehland2019-3}, \ie, if the model makes an inference about the system's behavior in a new range of variables \cite{Boehland2019-26}.
Böhland \etal \cite{Boehland2019} propose a hybrid model for local adaption of the model complexity to interpolation and extrapolation.
The hybrid model creates a hull around the interpolation areas using a fuzzy modeled One-Class Support Vector Machine (1C-SVM).
The fuzzified hull serves as a weighting function for the extrapolation and the interpolation model.

\subsection{Automated Design}
\label{ssec:AutomatedDesign}

The design algorithm for black-box modeling automatically determines a suitable combination of the interpolation and the extrapolation model (sub-models), as well as the 1C-SVM.
The data set is split into training, validation, and test data.
The algorithm creates candidate hybrid models using the training data and estimates their performance afterward with the validation data.
After selecting the best performing hybrid model, it is retrained using the training and validation data (tuning data), and the performance is assessed with the test data.

\paragraph{Grid Search}
\label{par:GridSearch}

The most elementary algorithm for optimizing the configuration $\boldsymbol{\lambda}$ of a model is grid search, where a finite set of candidate configurations is defined and exhaustively evaluated.
The configuration space $\boldsymbol{\Lambda}$ consists of sub-models of various prediction algorithms,
    incorporating the MultiLayer Perceptron (MLP),
    the SVR,
    Multivariate Adaptive Regression Splines (MARS) \cite{MARS},
    and the LOcal LInear MOdel Tree (LOLIMOT) \cite{LOLIMOT},
    and the 1C-SVM; each prediction and decision algorithm has a finite space of candidate hyperparameters.
The algorithm of Böhland \etal \cite{Boehland2019} creates the candidate models and gathers all trained sub-models and 1C-SVMs in the model pool.
Then, each possible combination of sub-models and 1C-SVMs of the pool is evaluated on the validation data, and the $\boldsymbol{\lambda}$ with the lowest MAE is selected.

\paragraph{Bayesian Optimization}
\label{par:BayesianOptimization}

Rather than evaluating a finite search grid, Bayesian optimization explores and exploits the configuration space $\boldsymbol{\Lambda}$.
The optimization scheme uses a probabilistic surrogate model to approximate the objective function $\mathcal{Q}$, mapping the model's performance $Q$ over $\boldsymbol{\Lambda}$.
In each iteration, the surrogate model is updated, and the optimization scheme uses an acquisition function to decide on the next hyperparameter configuration $\boldsymbol{\lambda} \in \boldsymbol{\Lambda}$ to be observed \cite{Feurer2019}.
To apply Bayesian optimization towards automated hybrid modeling, we need to define the configuration space $\boldsymbol{\Lambda}$.

\subsection{Evaluation}
\label{ssec:Evaluation}

We evaluate the automated design of the hybrid model as in reference \cite{Boehland2019} on ten data sets and compare grid search to Bayesian optimization.
\vspace{-5mm}
\paragraph{Experimental Setup}
\label{par:ExperimentalSetup}

In the initial proposal of the automated hybrid model, Böhland \etal \cite{Boehland2019} showed that it performs significantly better than standard regression models on nine out of ten benchmark data sets.
In this experiment, we compare HPO algorithms for automated model design, \ie, Bayesian optimization with exhaustive grid search.
We evaluate which HPO algorithm achieves lower prediction errors and requires less computation time.

For grid search, we adopt the configuration space $\boldsymbol{\Lambda}$ of reference \cite{Boehland2019} with the 1C-SVM implementation of the Scikit-learn library\footnote{\label{fn:sklearn}
    \url{https://github.com/scikit-learn/scikit-learn}
} \cite{sklearn} (RBF kernel; $\sigma =
    0.01,\allowbreak
    0.025,\allowbreak
    0.05,\allowbreak
    0.1,\allowbreak
    0.2,\allowbreak
    \ldots,\allowbreak
    1,\allowbreak
    1.5,\allowbreak
    10$; $\epsilon = 0.001$)\footnote{\label{EpsolonNu}some references denote $\epsilon$ as $\nu$}.
Since the LOLIMOT model is not available in the Python programming language \cite{Python}, we omit this model type but added Random Forest (RF), Gradient Boosting Machine (GBM), and Linear Regression (LR):
\begin{itemize}
    \small
    \item MLP ($N_\text{neurons} =
        2,\allowbreak
        3,\allowbreak
        \ldots,\allowbreak
        17,\allowbreak
        30,\allowbreak
        50$),
    \item SVR (RBF kernel; $\sigma =
        0.1,\allowbreak
        0.2,\allowbreak
        \ldots,\allowbreak
        1,\allowbreak
        1.2,\allowbreak
        1.5,\allowbreak
        2$; $C=100$; $\epsilon = 0.001$)\footnoteref{EpsolonNu},
    \item GBM ($N_\text{estimators} =
        90,\allowbreak
        100,\allowbreak
        \ldots,\allowbreak
        150,\allowbreak
        200,\allowbreak
        300,\allowbreak
        \ldots,\allowbreak
        1000$),
    \item RF ($N_\text{estimators} =
        90,\allowbreak
        100,\allowbreak
        \ldots,\allowbreak
        150,\allowbreak
        200,\allowbreak
        300,\allowbreak
        \ldots,\allowbreak
        1000$),
    \item MARS,
    \item LR
\end{itemize}
We implemented the automated design process in Python \cite{Python} and adapted the grid search from the implementation of the Scikit-learn library\footnoteref{fn:sklearn} \cite{sklearn}.
SVR, RF, and LR are based on the Scikit-learn library as well.
The GBM implementation is based on the XGBoost library\footnote{
    \url{https://github.com/dmlc/xgboost}
} \cite{XGboost},
    and MARS relies on the Py-earth library\footnote{
        \url{https://github.com/scikit-learn-contrib/py-earth}
    } \cite{Py-earth}.
For Bayesian optimization, we apply the NNI toolkit\footnote{
    \url{https://github.com/microsoft/nni}
} \cite{NNI} with the Tree Parzen Estimator (TPE) optimizer \cite{Hyperopt}.
The configuration space $\boldsymbol{\Lambda}$ consist of three categorical hyperparameters -- the choice for the interpolation, extrapolation, and decision algorithm.
For the interpolation and the extrapolation, the optimizer may choose the above-listed prediction algorithms.
If an algorithm is chosen, the corresponding conditional configuration space $\boldsymbol{\Lambda}_\text{cond} \subset \boldsymbol{\Lambda}$ with continuous values and the limits corresponding to the respective minimum and maximum values of the grid search applies.
For the decision algorithm, only the 1C-SVM algorithm with RBF kernel can be chosen with continuous hyperparameters and limits equivalent to the grid search.
\smallbreak
We evaluate the automated design process on ten data sets and split the data randomly into training data ($\SI{60}{\percent}$), validation data ($\SI{20}{\percent}$), and test data ($\SI{20}{\percent}$).
The test data is initially held out.
With the remaining data, we perform a four-fold CV for each candidate configuration $\boldsymbol{\lambda}$ and calculate the mean MAE over the splits.
Based on the mean MAE, the HPO algorithm determines the most suitable $\boldsymbol{\lambda}$. 
Afterward, the hybrid model is re-fitted with the chosen $\boldsymbol{\lambda}$ using the train and validation data (tuning data) and tested on the hold-out test data.
We repeat this process five times for each data set with different random seeds for splitting the data.

\paragraph{Results}
\label{par:Results}

We evaluate the performance of grid search and Bayesian optimization by comparing the MAE of the chosen configuration $\boldsymbol{\lambda}$ on the hold-out test data and the computation times.
\autoref{tab:hybrid-model-results} shows the experimental results regarding the prediction error MAE and the computing time.

\begin{table}[htbp]
	\centering
	\footnotesize
	\caption{Prediction error and computation time comparison of grid search and Bayesian optimization for the automated design of the hybrid model.}
	\begin{adjustbox}{max width=1\textwidth}
        \begin{tabular}{lrrrr}
        \toprule
        \textbf{Data Set} & \multicolumn{2}{c}{\textbf{Prediction Error MAE} [$10^{-3}$]} & \multicolumn{2}{c}{\textbf{Computation Time} [s]} \\
              & \multicolumn{1}{c}{grid search} & \multicolumn{1}{c}{Bayesian optimization} & \multicolumn{1}{c}{grid search} & \multicolumn{1}{c}{Bayesian optimization} \\
        \midrule
        Abalone & 52.98 & \textbf{52.81} & \textbf{111} & 3832 \\
        Airfoil & 27.70 & \textbf{26.72} & \textbf{51} & 2112 \\
        Boston & \textbf{49.87} & 50.22 & \textbf{39} & 1093 \\
        California & 64.99 & \textbf{64.78} & \textbf{607} & 12637 \\
        Computer & 16.85 & \textbf{16.56} & \textbf{387} & 7396 \\
        Concrete & 37.79 & \textbf{36.25} & \textbf{48} & 1176 \\
        Ailerons & \textbf{26.28} & 28.29 & \textbf{173} & 6049 \\
        Elevators & 40.50 & \textbf{40.29} & \textbf{261} & 8040 \\
        Redwine & \textbf{68.71} & 69.81 & \textbf{144} & 3541 \\
        Whitewine & \textbf{82.82} & 84.15 & \textbf{66} & 2091 \\
        \bottomrule
        \end{tabular}%
	\end{adjustbox}
	\label{tab:hybrid-model-results}%
\end{table}%

The comparison of the MAE on the hold-out test data shows that no HPO algorithm has a significant advantage in terms of prediction errors.
The advantage of Bayesian optimization is that only the boundaries of the configuration space $\boldsymbol{\Lambda}$ have to be defined.
Thus, configurations between the points of the grid search are considered, and a reasonable definition of $\boldsymbol{\Lambda}$ depends less on the skillful definition of candidates by the data scientist.

Grid search shows a clear advantage in terms of computing time.
The advantage can be justified with the re-usability of already trained sub-models.
For searching the configuration space $\boldsymbol{\Lambda}$ of the hybrid model, it is sufficient to fit the grid points of each prediction algorithm individually and cache the predictions on the validation data.
Then, we may calculate the prediction error for all possible $\boldsymbol{\lambda} \in \boldsymbol{\Lambda}$ combinations of the interpolator, extrapolator, and decider by combining the cached results (similar to \textit{dynamic programming}).
Thereby, the number of configurations to be assessed does not increase exponentially with the points of the grid search but linearly, resulting in $N_\text{MLP}$ + $N_\text{SVR}$ + $N_\text{MARS}$ + $N_\text{GBM}$ + $N_\text{RF}$ + $N_\text{LR}$ + $N_\text{1C-SVM} = 79$ evaluations.
In Bayesian optimization, in contrast, we define the number of candidate configurations (trials) to be evaluated $N_\text{trials} = 500$, and the optimization selects candidates based on the performance of previous trials (directed search).

\begin{figure}%
    \centering
    \pgfplotsset{width=\textwidth,height=5.5cm}
    \begin{tikzpicture}
    \tikzstyle{every node}=[font=\footnotesize]
    \begin{axis}[
        legend cell align={left},
        legend style={fill opacity=0.8, draw opacity=1, text opacity=1, draw=white!80!black},
        log basis x={10},
        x grid style={white!69.0196078431373!black},
        xlabel={Wall clock time in s},
        xmajorgrids,
        xmin=7.97847814612902, xmax=9339.37508322341,
        xmode=log,
        xtick style={color=black},
        xtick={0.1,1,10,100,1000,10000,100000},
        xticklabels={
          \(\displaystyle {10^{-1}}\),
          \(\displaystyle {10^{0}}\),
          \(\displaystyle {10^{1}}\),
          \(\displaystyle {10^{2}}\),
          \(\displaystyle {10^{3}}\),
          \(\displaystyle {10^{4}}\),
          \(\displaystyle {10^{5}}\)
        },
        y grid style={white!69.0196078431373!black},
        ylabel={Validation MAE},
        ymajorgrids,
        ymin=0.00213010960447145, ymax=0.0729498746959637,
        ytick style={color=black},
        font={\sffamily},
        legend style={cells={align=left},
        axis lines = box},
        tick align=inside,
    ]
    \path [draw=red, fill=red, opacity=0.2]
        (axis cs:11,0.021888388589612)
        --(axis cs:11,0.0697307944645323)
        --(axis cs:43,0.0697307944645323)
        --(axis cs:44,0.0631289891794899)
        --(axis cs:65,0.0631289891794899)
        --(axis cs:66,0.0631139509064584)
        --(axis cs:93,0.0631139509064584)
        --(axis cs:94,0.0500221547942152)
        --(axis cs:101,0.0500221547942152)
        --(axis cs:102,0.0498171131538876)
        --(axis cs:136,0.0498171131538876)
        --(axis cs:137,0.0261784852414908)
        --(axis cs:157,0.0261784852414908)
        --(axis cs:158,0.0261784847135319)
        --(axis cs:180,0.0261784847135319)
        --(axis cs:181,0.0261766254429405)
        --(axis cs:183,0.0261766254429405)
        --(axis cs:184,0.0260036413415907)
        --(axis cs:193,0.0260036413415907)
        --(axis cs:194,0.0253729716905286)
        --(axis cs:198,0.0253729716905286)
        --(axis cs:199,0.0200276736914812)
        --(axis cs:217,0.0200276736914812)
        --(axis cs:218,0.0200235143742716)
        --(axis cs:314,0.0200235143742716)
        --(axis cs:315,0.0174012200817859)
        --(axis cs:457,0.0174012200817859)
        --(axis cs:458,0.0173918232541003)
        --(axis cs:469,0.0173918232541003)
        --(axis cs:470,0.0173320986777705)
        --(axis cs:542,0.0173320986777705)
        --(axis cs:543,0.0172797848388772)
        --(axis cs:579,0.0172797848388772)
        --(axis cs:580,0.0172633297883782)
        --(axis cs:665,0.0172633297883782)
        --(axis cs:666,0.0172621816092732)
        --(axis cs:704,0.0172621816092732)
        --(axis cs:705,0.0172592822363591)
        --(axis cs:772,0.0172592822363591)
        --(axis cs:773,0.0172588483220862)
        --(axis cs:941,0.0172588483220862)
        --(axis cs:942,0.0172573319213122)
        --(axis cs:1011,0.0172573319213122)
        --(axis cs:1012,0.0172567609778067)
        --(axis cs:1053,0.0172567609778067)
        --(axis cs:1054,0.0172530867402343)
        --(axis cs:1191,0.0172530867402343)
        --(axis cs:1192,0.0172449131957714)
        --(axis cs:1194,0.0172449131957714)
        --(axis cs:1195,0.0172447046679277)
        --(axis cs:1240,0.0172447046679277)
        --(axis cs:1241,0.0172429213443399)
        --(axis cs:1265,0.0172429213443399)
        --(axis cs:1266,0.0172429198381857)
        --(axis cs:2122,0.0172429198381857)
        --(axis cs:2123,0.0172369774447069)
        --(axis cs:2166,0.0172369774447069)
        --(axis cs:2167,0.0172368237752038)
        --(axis cs:2541,0.0172368237752038)
        --(axis cs:2542,0.0172365716098948)
        --(axis cs:2557,0.0172365716098948)
        --(axis cs:2558,0.0172364621042182)
        --(axis cs:2770,0.0172364621042182)
        --(axis cs:2771,0.0172358536087042)
        --(axis cs:3501,0.0172358536087042)
        --(axis cs:3502,0.0172357197459644)
        --(axis cs:3719,0.0172357197459644)
        --(axis cs:3720,0.0172353661958921)
        --(axis cs:3858,0.0172353661958921)
        --(axis cs:3859,0.0172289504424336)
        --(axis cs:4423,0.0172289504424336)
        --(axis cs:4424,0.0172283879625897)
        --(axis cs:5481,0.0172283879625897)
        --(axis cs:5482,0.0172276583344577)
        --(axis cs:5937,0.0172276583344577)
        --(axis cs:5938,0.0172275189247859)
        --(axis cs:6069,0.0172275189247859)
        --(axis cs:6070,0.017227465434749)
        --(axis cs:6191,0.017227465434749)
        --(axis cs:6192,0.0172274435548027)
        --(axis cs:6312,0.0172274435548027)
        --(axis cs:6313,0.0172190397371121)
        --(axis cs:6454,0.0172190397371121)
        --(axis cs:6455,0.0172010583098239)
        --(axis cs:6773,0.0172010583098239)
        --(axis cs:6774,0.0172010309656256)
        --(axis cs:6774,0.0168178213797391)
        --(axis cs:6774,0.0168178213797391)
        --(axis cs:6773,0.0168181410573439)
        --(axis cs:6455,0.0168181410573439)
        --(axis cs:6454,0.0168114534074186)
        --(axis cs:6313,0.0168114534074186)
        --(axis cs:6312,0.0168082338364074)
        --(axis cs:6192,0.0168082338364074)
        --(axis cs:6191,0.0168084855344643)
        --(axis cs:6070,0.0168084855344643)
        --(axis cs:6069,0.0168090704076594)
        --(axis cs:5938,0.0168090704076594)
        --(axis cs:5937,0.0168104333857138)
        --(axis cs:5482,0.0168104333857138)
        --(axis cs:5481,0.0168111416136774)
        --(axis cs:4424,0.0168111416136774)
        --(axis cs:4423,0.0168113492230297)
        --(axis cs:3859,0.0168113492230297)
        --(axis cs:3858,0.0168088715550403)
        --(axis cs:3720,0.0168088715550403)
        --(axis cs:3719,0.0168092163129086)
        --(axis cs:3502,0.0168092163129086)
        --(axis cs:3501,0.0168093445996916)
        --(axis cs:2771,0.0168093445996916)
        --(axis cs:2770,0.0168101307217369)
        --(axis cs:2558,0.0168101307217369)
        --(axis cs:2557,0.0168100881062603)
        --(axis cs:2542,0.0168100881062603)
        --(axis cs:2541,0.016810188040205)
        --(axis cs:2167,0.016810188040205)
        --(axis cs:2166,0.01681033622601)
        --(axis cs:2123,0.01681033622601)
        --(axis cs:2122,0.0168080158871375)
        --(axis cs:1266,0.0168080158871375)
        --(axis cs:1265,0.0168080316846834)
        --(axis cs:1241,0.0168080316846834)
        --(axis cs:1240,0.0168087402892264)
        --(axis cs:1195,0.0168087402892264)
        --(axis cs:1194,0.0168089500545019)
        --(axis cs:1192,0.0168089500545019)
        --(axis cs:1191,0.0168072467282062)
        --(axis cs:1054,0.0168072467282062)
        --(axis cs:1053,0.0168106781294116)
        --(axis cs:1012,0.0168106781294116)
        --(axis cs:1011,0.0168111410528412)
        --(axis cs:942,0.0168111410528412)
        --(axis cs:941,0.0168105524917572)
        --(axis cs:773,0.0168105524917572)
        --(axis cs:772,0.0168154894394801)
        --(axis cs:705,0.0168154894394801)
        --(axis cs:704,0.0168143600890958)
        --(axis cs:666,0.0168143600890958)
        --(axis cs:665,0.0168216749379762)
        --(axis cs:580,0.0168216749379762)
        --(axis cs:579,0.0168151832175292)
        --(axis cs:543,0.0168151832175292)
        --(axis cs:542,0.0168427996984406)
        --(axis cs:470,0.0168427996984406)
        --(axis cs:469,0.0168223787510176)
        --(axis cs:458,0.0168223787510176)
        --(axis cs:457,0.0168424244197506)
        --(axis cs:315,0.0168424244197506)
        --(axis cs:314,0.0157079897378428)
        --(axis cs:218,0.0157079897378428)
        --(axis cs:217,0.0157570474230183)
        --(axis cs:199,0.0157570474230183)
        --(axis cs:198,0.0132470285498904)
        --(axis cs:194,0.0132470285498904)
        --(axis cs:193,0.0140666791552318)
        --(axis cs:184,0.0140666791552318)
        --(axis cs:183,0.0139901125379237)
        --(axis cs:181,0.0139901125379237)
        --(axis cs:180,0.0140526020882528)
        --(axis cs:158,0.0140526020882528)
        --(axis cs:157,0.0140526936586398)
        --(axis cs:137,0.0140526936586398)
        --(axis cs:136,0.00534918983590292)
        --(axis cs:102,0.00534918983590292)
        --(axis cs:101,0.00698466356573218)
        --(axis cs:94,0.00698466356573218)
        --(axis cs:93,0.0129169840689548)
        --(axis cs:66,0.0129169840689548)
        --(axis cs:65,0.0129136620019492)
        --(axis cs:44,0.0129136620019492)
        --(axis cs:43,0.021888388589612)
        --(axis cs:11,0.021888388589612)
        --cycle;
    
    \addplot [thick, red]
        table {%
        11 0.0458095915270721
        43 0.0458095915270721
        44 0.0380213255907195
        65 0.0380213255907195
        66 0.0380154674877066
        93 0.0380154674877066
        94 0.0285034091799737
        101 0.0285034091799737
        102 0.0275831514948952
        136 0.0275831514948952
        137 0.0201155894500653
        157 0.0201155894500653
        158 0.0201155434008924
        180 0.0201155434008924
        181 0.0200833689904321
        183 0.0200833689904321
        184 0.0200351602484112
        193 0.0200351602484112
        194 0.0193100001202095
        198 0.0193100001202095
        199 0.0178923605572498
        217 0.0178923605572498
        218 0.0178657520560572
        314 0.0178657520560572
        315 0.0171218222507683
        457 0.0171218222507683
        458 0.017107101002559
        469 0.017107101002559
        470 0.0170874491881056
        542 0.0170874491881056
        543 0.0170474840282032
        579 0.0170474840282032
        580 0.0170425023631772
        665 0.0170425023631772
        666 0.0170382708491845
        704 0.0170382708491845
        705 0.0170373858379196
        772 0.0170373858379196
        773 0.0170347004069217
        941 0.0170347004069217
        942 0.0170342364870767
        1011 0.0170342364870767
        1012 0.0170337195536091
        1053 0.0170337195536091
        1054 0.0170301667342203
        1191 0.0170301667342203
        1192 0.0170269316251367
        1194 0.0170269316251367
        1195 0.0170267224785771
        1240 0.0170267224785771
        1241 0.0170254765145117
        1265 0.0170254765145117
        1266 0.0170254678626616
        2122 0.0170254678626616
        2123 0.0170236568353585
        2166 0.0170236568353585
        2167 0.0170235059077044
        2541 0.0170235059077044
        2542 0.0170233298580776
        2557 0.0170233298580776
        2558 0.0170232964129776
        2770 0.0170232964129776
        2771 0.0170225991041979
        3501 0.0170225991041979
        3502 0.0170224680294365
        3719 0.0170224680294365
        3720 0.0170221188754662
        3858 0.0170221188754662
        3859 0.0170201498327317
        4423 0.0170201498327317
        4424 0.0170197647881336
        5481 0.0170197647881336
        5482 0.0170190458600858
        5937 0.0170190458600858
        5938 0.0170182946662226
        6069 0.0170182946662226
        6070 0.0170179754846066
        6191 0.0170179754846066
        6192 0.0170178386956051
        6312 0.0170178386956051
        6313 0.0170152465722654
        6454 0.0170152465722654
        6455 0.0170095996835839
        6773 0.0170095996835839
        6774 0.0170094261726823
    };
    \addplot [scatter, only marks, draw=blue, draw=none, fill=blue, mark=*]
    table{%
        x  y
        386.936533021927 0.0170921262160324
    };
    \addlegendentry{Bayesian optimization}
    \addlegendentry{Grid search}
    \end{axis}
    \end{tikzpicture}
    \caption{
        Development of the MAE on the validation splits of the \textit{computer} data set using Bayesian optimization and grid search.
        The HPO was repeated five times with random splits.
        The red line reflects the mean progress over the independent runs, and the light red area the $\SI{95}{\percent}$ confidence interval, determined based on the Student's t-distribution. 
    }
    \label{fig:convergence-MAE-computer}%
\end{figure}
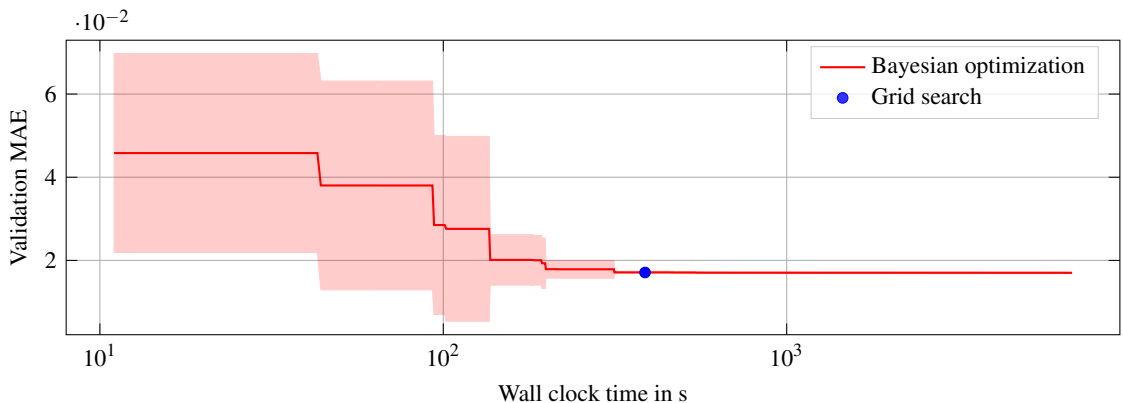

\autoref{fig:convergence-MAE-computer} shows the evolution of the mean MAE on the validation splits of the \textit{computer} data set using Bayesian optimization and grid search.
The $\SI{95}{\percent}$ confidence interval was determined over the five loops based on the Student's t-distribution.
The progression of the Bayesian optimization converges well before the $\nth{500}$ trial.
Thus, there is the potential of terminating the Bayesian optimization prematurely as soon as we are satisfied with the result.
In contrast, the grid search cannot be terminated prematurely, as otherwise, areas of the configuration space $\boldsymbol{\Lambda}$ would not have been examined, and we would obtain an incomplete model.
Therefore, it is reasonable to implement convergence regularization in future work that terminates the optimization automatically if no further improvement is expected (\textit{early stopping}).

In future work, we plan to integrate the hybrid model into pyWATTS \cite{Heidrich2021} with the grid search.
The convergence regularization will be developed for the pyWATTS HPO extension (see \autoref{ssec:HPOExtension}), for HPO problems with exponential complexity.
\\
In addition to the application as an EV battery model for representing the electrical behavior shown in reference \cite{Schwenk2021}, we target to model thermal building dynamics, using black-box models for interpolation and grey-box models (thermal-electrical analogy) for extrapolation.
In this way, we target to design a robust default template for thermal building modeling, which can be used for demand side management, \eg, using model predictive control.

\section{Conclusion and Outlook}
\label{sec:conclusion_outlook}

The transformation of a fossil-based to a sustainable energy system requires the large-scale application of machine learning algorithms.
For satisfying the rapidly growing demand for time series forecasts, we need to automate the design and application.
We proposed five automation levels, where Automation level 0 is manual design and application and Automation level 5 is an autonomous system.
For Automation levels 1, 2, 3, and 4, we introduced forecasting approaches for smart grid applications and described their concepts.
\smallskip\\
For one of the approaches -- the hybrid model -- we evaluated two HyperParameter Optimization (HPO) algorithms for the automated design (Automation level 2.5).
The hybrid model weights the results of two models depending on whether the input values were represented in the training data set or not.
In this way, a robust model is used for extrapolation and a sophisticated model for interpolation.
The evaluation shows an advantage of grid search in terms of computation time if we re-use already trained models.
Regarding the prediction error, there is no clear advantage of grid search or Bayesian optimization.
\smallskip\\
In future work, a performance tracking and warning system could monitor the templates' operation and alert the data scientist if degrading forecasting performance is detected.
We target to improve the PhotoVoltaic (PV) and the Wind Power (WP) forecasting templates by online calibration (Automation level 3).
More precisely, the PV template will be calibrated for individual plant efficiencies, orientations, and alignments, and the WP template will include HPO for the friction coefficient $\alpha$.
For Automation level 4, we develop the \textit{Pipeline Wizard}, automating the design of the complete forecasting pipeline.
The \textit{Pipeline Wizard} includes automated data selection or generation, online performance prediction, and adaption strategies for concept drifts.
In the long view, probabilistic interval forecasts will replace point forecasts, \eg, reference \cite{Phipps2021}.
\smallskip\\
We plan to integrate the proposed automation approaches in the Energy Lab 2.0 \cite{EnergyLab2}, a real-world research environment for exploring intelligent coupling of various energy generation, storage, and supply capabilities.
The approaches for each automation level help to solve forecasting tasks according to individual complexity and requirements.

\section*{Acknowledgements}

This project is funded by the Helmholtz Association’s Initiative and Networking Fund through \textit{Helmholtz AI}, the Helmholtz Association under the Program \textit{Energy System Design}.
The authors would like to thank \textit{Stadtwerke Karlsruhe Netzservice GmbH} (Karlsruhe, Germany) for the provided PV and WP data, and Moritz Böhland for sharing the Matlab implementation of the Hybrid Model \cite{Boehland2019}.
Conceptualization and methodology: S.~M., R.~M.;
Experiments, validation, investigation, and visualization: J.~P., T.~M., S.~M.;
Writing -- original draft preparation: S.~M.;
Writing -- review and editing: S.~M., J.~P., T.~M., V.~H., R.~M.;
Supervision and funding acquisition: V.~H., R.~M.;
All authors have read and agreed to the published version of the article.

\bibliographystyle{ieeetr}

\begin{thebibliography}{99}
    \small
    \bibitem{Ahmed2020}
    R.~Ahmed, V.~Sreeram, Y.~Mishra, and M.~D.~Arif,
    ``A review and evaluation of the state-of-the-art in PV solar power forecasting: Techniques and optimization,''
    in Renewable and Sustainable Energy Reviews,
    vol.~124,
    pp.~109792,
    2020.
    \bibitem{Qian2019}
    Z.~Qian, Y.~Pei, H.~Zareipour, and N.~Chen,
    ``A review and discussion of decomposition-based hybrid models for wind energy forecasting applications,''
    in Applied Energy,
    vol.~235,
    pp.~939--953,
    2019.
    \bibitem{Nti2020}
    I.~K.~Nti, M.~Teimeh, O.~Nyarko-Boateng, and A.~F.~Adekoya,
    ``Electricity load forecasting: A systematic review,''
    in Journal of Electrical Systems and Information Technology,
    vol.~7,
    no.~13,
    2020.
    \bibitem{Fridgen2020}
    G.~Fridgen, R.~Keller, M.~F.~Körner, and M.~Schöpf,
    ``A holistic view on sector coupling,''
    in Energy Policy,
    vol.~147,
    pp.~111913,
    2020.
    \bibitem{SAE-J3016}
    Society of Automotive Engineers,
    ``SAE J3016 levels of driving automation,''
    2021,
    \url{https://www.sae.org/standards/content/j3016_202104/}.
    \bibitem{Heidrich2021}
    B.~Heidrich, A.~Bartschat, M.~Turowski, O.~Neumann, K.~Phipps, S.~Meisenbacher, K.~Schmieder, N.~Ludwig, R.~Mikut, and V.~Hagenmeyer,
    ``pyWATTS: Python Workflow Automation Tool for Time Series,''
    arXiv:~2106.10157,
    2021.
    \bibitem{GonzalezOrdiano2018}
    J.~A.~González Ordiano, S.~Waczowicz, V.~Hagenmeyer, and R.~Mikut,
    ``Energy forecasting tools and services,''
    in WIREs Data Mining and Knowledge Discovery,
    vol.~8,
    no.~2,
    pp.~e1235,
    2018.
    \bibitem{Hyperopt}
    J.~Bergstra, D.~Yamins, and D.~D.~Cox,
    ``Making a science of model search: Hyperparameter optimization in hundreds of dimensions for vision architectures,''
    in Proceedings of the 30. International Conference on Machine Learning,
    pp.~I-115--I-123,
    2013,
    \url{http://hyperopt.github.io/hyperopt/}.
    \bibitem{SMAC3}
    M.~Lindauer, K.~Eggensperger, M.~Feurer, S.~Falkner, A.~Biedenkapp, and F.~Hutter,
    ``SMAC v3: Algorithm configuration in Python,'
    2017,
    \url{https://automl.github.io/SMAC3/master/}.
    \bibitem{NNI}
    Microsoft Corporation,
    ``Neural Network Intelligence,''
    2017,
    \url{https://nni.readthedocs.io/en/stable/index.html}.
    \bibitem{Ratz2019}
    M.~Rätz, A.~P.~Javadi, M.~Baranski, K.~Finkbeiner, and D.~Müller,
    ``Automated data-driven modeling of building energy systems via machine learning algorithms,''
    in Energy and Buildings,
    vol.~202,
    pp.~109384,
    2019.
    \bibitem{Cui2016}
    C.~Cui, T.~Wu, M.~Hu, J.~D.~Weir, and X.~Li,
    ``Short-term building energy model recommendation system: A meta-learning approach,''
    in Applied Energy,
    vol.~172,
    pp.~251--263,
    2016.
    \bibitem{Shahoud2020a}
    S.~Shahoud, H.~Khalloof, M.~Winter, C.~Duepmeier, and V.~Hagenmeyer,
    ``A meta learning approach for automating model selection in big data environments using microservice and container virtualization technologies,''
    in Proceedings of the 12. International Conference on Management of Digital EcoSystems,
    pp.~84--91,
    2020.
    \bibitem{Wu2020}
    Z.~Wu, X.~Xia, L.~Xiao, and Y.~Liu,
    ``Combined model with secondary decomposition-model selection and sample selection for multi-step wind power forecasting,''
    in Applied Energy,
    vol.~261,
    pp.~114345,
    2020.
    \bibitem{Maldonado2019}
    S.~Maldonado, A.~González, and S.~Crone,
    ``Automatic time series analysis for electric load forecasting via support vector regression,''
    in Applied Soft Computing,
    vol.~83,
    pp.~105616,
    2019.
    \bibitem{Valente2020}
    J.~M.~Valente, and S.~Maldonado,
    ``SVR-FFS: A novel forward feature selection approach for high-frequency time series forecasting using support vector regression,''
    in Expert Systems with Applications,
    vol.~160,
    pp.~113729,
    2020.
    \bibitem{Schachinger2018}
    D. Schachinger, J. Pannosch, and W. Kastner,
    ``Adaptive learning-based time series prediction framework for building energy management,''
    in Proceedings of the 2018 IEEE International Conference on Industrial Electronics for Sustainable Energy Systems (IESES),
    pp.~453--458,
    2018.
    \bibitem{AutoSklearn}
    M.~Feurer, A.~Klein, K.~Eggensperger, J.~Springenberg, M.~Blum, and F.~Hutter,
    ``Efficient and robust automated machine learning,''
    in Advances in Neural Information Processing Systems,
    vol.~28,
    pp.~2962--2970,
    2015,
    \url{https://automl.github.io/auto-sklearn/master/}.
    \bibitem{TPOT}
    T.~T.~Le, W.~Fu, and J.~H.~Moore,
    ``Scaling tree-based automated machine learning to biomedical big data with a feature set selector,''
    in Bioinformatics,
    vol.~36,
    no.~1,
    pp.~250--256,
    2020,
    \url{http://epistasislab.github.io/tpot/}.
    \bibitem{Zoeller2021}
    M.-A.~Zöller, and M.~Huber,
    ``Benchmark and survey of automated machine learning frameworks,''
    in Journal Artificial Intelligence Research,
    vol.~70,
    pp.~409--472,
    2021.
    \bibitem{Santu2021}
    S.~K.~K.~Santu, M.~M.~Hassan, M.~J.~Smith, L.~Xu, C.~Zhai, and K.~Veeramachaneni,
    ``A level-wise taxonomic perspective on automated machine learning to date and beyond: Challenges and opportunities,''
    arXiv:~2010.10777,
    2021.
    \bibitem{Chaouachi2010}
    A.~Chaouachi, R.~M.~Kamel, and K.~Nagasaka,
    ``Neural network ensemble-based solar power generation short-term forecasting,''
    in Journal of Advanced Computational Intelligence and Intelligent Informatics,
    vol.~14,
    no.~1,
    pp.~69--75,
    2010.
    \bibitem{Mellit2010}
    A.~Mellit, and A.~M.~Pavan,
    ``A 24-h forecast of solar irradiance using artificial neural network: Application for performance prediction of a grid-connected PV plant at Trieste, Italy,''
    in Solar Energy,
    vol.~84,
    no.~5,
    pp.~807--821,
    2010.
    \bibitem{Atique2019}
    S.~Atique, S.~Noureen, V.~Roy, V. ~Subburaj, S.~Bayne, and J.~Macfie,
    ``Forecasting of total daily solar energy generation using ARIMA: A case study,''
    in Proceedings of the IEEE 9. Annual Computing and Communication Workshop and Conference (CCWC),
    pp.~0114--0119,
    2019.
    \bibitem{GridSage}
    Zentrum für Sonnenenergie- und Wasserstoff-Forschung Baden-Württemberg (ZSW),
    ``GridSage,''
    accessed: 27.08.2021,
    2021,
    \url{https://www.zsw-bw.de/leistung/netzintegration-und-mobilitaet/gridsage-prognosen-fuer-den-redispatch-20.html}.
    \bibitem{FuturePowerFlow}
    energy \& meteo systems GmbH,
    ``FuturePowerFlow,''
    accessed: 27.08.2021,
    \url{https://www.emsysgrid.de/produkte/redispatch/futurepowerflow.php}.
    \bibitem{Redispatching}
    KISTERS AG,
    ``KISTERS Redispatching 2.0,''
    accessed: 27.08.2021,
    \url{https://www.redispatching.de}.
    \bibitem{Martin2021}
    T.~Martin,
    ``Redispatch 2.0: Prognose der Einspeisemenge erneuerbarer Energien mithilfe von Machine Learning Ansätzen,''
    Bachelor's thesis,
    Karlsruhe Institute of Technology,
    Institute for Automation and Applied Informatics,
    2021.
    \bibitem{Masters2004}
    G.~M.~Masters,
    ``Renewable and efficient electric power systems,''
    1. ed.,
    Wiley,
    New York,
    2004.
    \bibitem{Weber2021}
    M.~Weber, M.~Turowski, H.~K.~Çakmak, R.~Mikut, U.~Kühnapfel, and V.~Hagenmeyer,
    ``Data-driven copy-paste imputation for energy time series,"
    in IEEE Transactions on Smart Grid,
    early access,
    2021.
    \bibitem{Zliobaite2016}
    I.~Žliobaitė, M.~Pechenizkiy, and J.~Gama,
    ``An overview of concept drift applications,''
    in N.~Japkowicz, J.~Stefanowski (eds) Big Data Analysis: New Algorithms for a New Society,
    Studies in Big Data,
    vol.~16,
    pp.~91--114,
    Springer,
    Cham,
    2016.
    \bibitem{Gama2014}
    J.~Gama, I.~Žliobaitė, A.~Bifet, M.~Pechenizkiy, and A.~Bouchachia,
    ``A survey on concept drift adaptation,''
    in ACM Computing Surveys,
    vol.~46,
    n.~4,
    2014.
    \bibitem{Celik2021}
    B.~Celik, and J.~Vanschoren,
    ``Adaptation strategies for automated machine learning on evolving data,''
    in IEEE Transactions on Pattern Analysis and Machine Intelligence,
    vol.~43,
    no.~9,
    pp.~3067--3078,
    2021.
    \bibitem{Schwenk2021}
    K.~Schwenk, S.~Meisenbacher, B.~Briegel, T.~Harr, V.~Hagenmeyer, and R.~Mikut,
    ``Integrating battery aging in the optimization for bidirectional charging of electric vehicles,''
    in IEEE Transactions on Smart Grid,
    early access,
    2021.
    \bibitem{Boehland2019-3}
    M.~A.~Babyak,
    ``What you see may not be what you get: A brief, nontechnical introduction to overfitting in regression-type models,''
    in Psychosomatic Medicine,
    vol.~66,
    no.~3,
    pp.~411--421,
    2004.
    \bibitem{Boehland2019-26}
    N.~Matalas, and V.~Bier,
    ``B. Extremes, extrapolation, and surprise,''
    in Risk Analysis,
    vol.~19,
    no.~1,
    pp.~49--54,
    1999.
    \bibitem{Boehland2019}
    M.~Böhland, W.~Doneit, L.~Gröll, R.~Mikut, and M.~Reischl,
    ``Automated design process for hybrid regression modeling with a one-class SVM,''
    in at - Automatisierungstechnik,
    vol.~67,
    no.~10,
    pp.~843--852,
    2019.
    \bibitem{MARS}
    J.~H.~Friedman,
    ``Multivariate adaptive regression splines,''
    in The Annals of Statistics,
    vol.~19,
    no.~1,
    pp.~1--67,
    1991.
    \bibitem{LOLIMOT}
    B.~Hartmann, T.~Ebert, T.~Fischer, J.~Belz, G.~Kampmann, and O.~Nelles,
    ``LMNTOOL–Toolbox zum Automatischen Trainieren Lokaler Modellnetze,''
    in Proceedings of the 22. Workshop Computational Intelligence,
    pp.~341--355,
    2014.
    \bibitem{Feurer2019}
    M.~Feurer, and F.~Hutter,
    ``Hyperparameter optimization,''
    in F.~Hutter, L.~Kotthoff, J.~Vanschoren (eds) Automated Machine Learning: Methods, Systems, Challenges,
    pp.~3--33,
    Springer International Publishing,
    Cham,
    2019.
    \bibitem{sklearn}
    F.~Pedregosa, G.~Varoquaux, A.~Gramfort, V.~Michel, B.~Thirion, O.~Grisel, M.~Blondel, P.~Prettenhofer, R.~Weiss, V.~Dubourg, J.~Vanderplas, A.~Passos, D.~Cournapeau, M.~Brucher, M.~Perrot, and E.~Duchesnay,
    ``Scikit-learn: Machine learning in Python,''
    in Journal of Machine Learning Research,
    vol.~12,
    pp.~2825--2830,
    2011,
    \url{https://scikit-learn.org/stable/index.html}.
    \bibitem{Python}
    Python Software Foundation,
    ``Python language reference, version 3.8,''
    \url{http://www.python.org}.
    \bibitem{XGboost}
    Chen, Tianqi, and C.~Guestrin,
    ``XGBoost: A scalable tree boosting system,''
    in Proceedings of the 22. ACM SIGKDD International Conference on Knowledge Discovery and Data Mining,
    pp.~785--794,
    2016,
    \url{https://xgboost.readthedocs.io/en/latest/index.html}.
    \bibitem{Py-earth}
    J.~Rudy,
    ``Py-earth,''
    2013,
    \url{https://contrib.scikit-learn.org/py-earth/}.
    \bibitem{Phipps2021}
    K.~Phipps, S.~Lerch, M.~Andersson, R.~Mikut, V.~Hagenmeyer, and N.~Ludwig,
    ``Evaluating ensemble post-processing for wind power forecasts,''
    arXiv:~2009.14127,
    2021.
    \bibitem{EnergyLab2}
    V.~Hagenmeyer, K.~C.~Hüseyin, C.~Düpmeier, T.~Faulwasser, J.~Isele, H.~B.~Keller, P.~Kohlhepp, U.~Kühnapfel, U.~Stucky, S.~Waczowicz, and R.~Mikut,
    ``Information and communication technology in Energy Lab 2.0: Smart Energies System Simulation and Control Center with an open-street-map-based power flow simulation example,''
    in Energy Technology,
    vol.~4,
    no.~1,
    pp.~145--162,
    2016.

\end{thebibliography}


\end{document}